\documentclass[conference]{IEEEtran}
\IEEEoverridecommandlockouts

\usepackage{newtxtext}   
\usepackage{newtxmath}   
\usepackage{amsmath}      
\usepackage[protrusion=false]{microtype}  
\usepackage{graphicx}
\usepackage{float}
\usepackage{booktabs}
\usepackage{array}       
\usepackage{siunitx}
\usepackage{xcolor}
\usepackage{balance}
\usepackage{cuted}
\usepackage{etoolbox}
\makeatletter\@ifundefined{@setmarks}{\let\@setmarks\relax}{}\makeatother
\usepackage{capt-of}
\usepackage[hidelinks]{hyperref}

\newcommand{\bodytablecaption}[1]{%
  \refstepcounter{table}%
  \centering\footnotesize\normalfont TABLE~\Roman{table}.\enspace #1\par
}
\makeatletter
\renewcommand{\section}{\@startsection{section}{1}{\z@}%
  {1.2ex plus .4ex minus .2ex}{0.8ex plus .2ex}%
  {\normalfont\normalsize\centering\uppercase}}
\renewcommand{\subsection}[1]{\@startsection{subsection}{2}{\z@}%
  {1.0ex plus .3ex minus .2ex}{0.5ex plus .2ex}%
  {\normalfont\normalsize\itshape\raggedright}{#1}}
\makeatother
\makeatletter
\apptocmd{\thebibliography}{%
  \fontsize{7.4}{8.05}\selectfont
  \setlength{\itemsep}{0pt}\setlength{\parsep}{0pt}\setlength{\parskip}{0pt}%
  \sloppy
}{}{}
\makeatother

\begin{document}

\title{\fontsize{20}{24}\selectfont Contact-Adaptive Robotic Ultrasound Probe Control for
Tissue Exploration and Continuous Task-Relevant Visualization Using Robot-Free
Image--Motion Demonstration}

\author{%
Seong Jeong\textsuperscript{*,1,2,3}, Minsung Kim\textsuperscript{1,2,4},
Dongho Yee\textsuperscript{1,2,4,5}, Juahn Oh\textsuperscript{1,2,8},
Yechan Seo\textsuperscript{1,2,3}, Jiyul Lee\textsuperscript{1,2,3},\\
Jinseok Lee\textsuperscript{2,4}, Seonho Shim\textsuperscript{2,9},
Younghoon Noh\textsuperscript{2,4}, Youngbin Kong\textsuperscript{1,7}
and Hyoun-Joong Kong\textsuperscript{\dag,1,3,6}
\thanks{\textsuperscript{*}First author. \textsuperscript{\dag}Corresponding author.}
\thanks{\textsuperscript{1}Department of Transdisciplinary Medicine, Seoul National University
Hospital. \textsuperscript{2}Rosota Inc., Seoul, Republic of Korea.
\textsuperscript{3}Department of Medicine, Seoul National University College of Medicine.
\textsuperscript{4}Department of Mechanical Engineering, Seoul National University.
\textsuperscript{5}Department of Computer Science and Engineering, Seoul National University.
\textsuperscript{6}Institute of Convergence Medicine with Innovative Technology, Seoul National
University Hospital. \textsuperscript{7}Interdisciplinary Program in Medical Informatics, Seoul
National University College of Medicine. \textsuperscript{8}Eulji University College of Medicine,
Daejeon, Republic of Korea. \textsuperscript{9}Department of Mechanical Engineering, Chungang
University, Seoul, Republic of Korea.}}

\IEEEpubid{\raisebox{-13pt}[0pt][0pt]{\parbox{\textwidth}{\scriptsize This work has been submitted
to the IEEE for possible publication. Copyright may be transferred without notice, after which this
version may no longer be accessible.}}}

\hypersetup{pdfauthor={Seong Jeong, Minsung Kim, Dongho Yee, Juahn Oh, Yechan Seo, Jiyul Lee, Jinseok Lee, Seonho Shim, Younghoon Noh, Youngbin Kong, Hyoun-Joong Kong},
            pdftitle={Contact-Adaptive Robotic Ultrasound Probe Control for Tissue Exploration and Continuous Task-Relevant Visualization Using Robot-Free Image-Motion Demonstration},
            pdfsubject={Robotic ultrasound; robot-free image--motion demonstrations; contact-adaptive probe control},
            pdfkeywords={robotic ultrasound, medical robotics, learning from demonstration, force regulation, HoLEP}}

\maketitle

\begin{strip}
\vspace{-2.55cm}
\centering\noindent
\includegraphics[width=\textwidth]{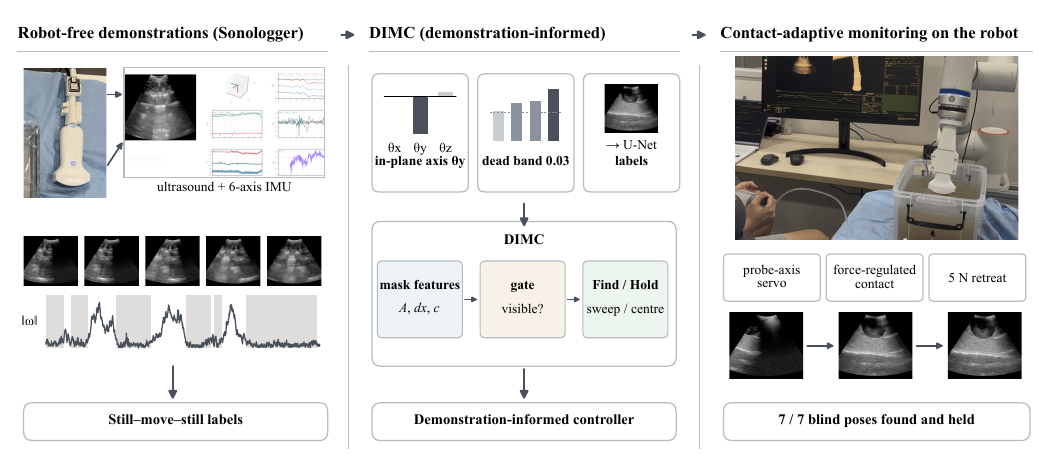}
{\setlength{\abovecaptionskip}{-2pt}%
\captionof{figure}{\textbf{Overview of the framework.} Robot-free image--motion
demonstrations recorded with Sonologger from the clinician-operated probe
(left) are converted into the parameters of the Demonstration-Informed
Image--Motion Controller (DIMC): the corrective probe axis, its sign and its
activation threshold, together with perception labels (middle). DIMC is
combined with beam-axis force regulation and an independent safety supervisor
for contact-adaptive robotic ultrasound monitoring, validated blind on a
dynamic phantom whose lumen volume and contact condition are changed by
syringe (right).}
\label{fig:overview}
}
\end{strip}

\begin{abstract}
Robotic ultrasound commonly targets standardized views, predefined scanning
protocols, or expert-scan reproduction. We target a different role: while a
clinician performs the primary procedure, a robotic assistant maintains a
clinician-selected view so that changing tissue remains observable. We present
contact-adaptive robotic ultrasound monitoring derived from robot-free
demonstrations. Sonologger clamps a six-axis inertial sensor to the
clinician-operated probe and records synchronized ultrasound and probe-frame
relative rotations without a robot, camera, or tracker. Rather than imitating
trajectories, the recordings measure which probe axis and sign the clinician
uses to correct an image offset. A measured image latency is removed before
still--move--still segmentation yields relative-rotation labels; translation
is excluded from control because of inertial drift. These measurements parameterize the
Demonstration-Informed Image--Motion Controller (DIMC), an interpretable,
non-learned controller combined with beam-axis force regulation and an
independent safety supervisor with \SI{5}{\newton} forced retreat. On a dynamic
bladder phantom, DIMC acquired and held a task-relevant view in 7 of 7 blind,
pose-paired trials versus 1 of 7 for a static hold (McNemar $p =
\num{0.031}$), and retained the view during syringe-driven hydro-distension
while the supervisor bounded the force excursion. The framework offers an
operational representation of ultrasound context for contact-adaptive
intraoperative monitoring; clinical validation remains future work.
\end{abstract}

\section{Introduction}
\IEEEpubidadjcol

Much robotic ultrasound research targets standardized examinations,
predefined views, or expert scanning behavior~\cite{li2021overview,jiang2023robotic}.
We target a complementary role: a safety-monitoring assistant that maintains
a clinician-selected view while the primary procedure continues. Intraoperative
ultrasound is a clinical localization adjunct in robot-assisted partial
nephrectomy~\cite{dicosmo2018intraoperative}; it motivates, but does not
clinically validate, the phantom study here. Maintaining the view is difficult
because tissue geometry and probe contact change together.

Conventional controller demonstrations are often collected on the robot.
Sonologger instead records ultrasound and probe-frame relative rotation from a
clinician-operated probe on one clock, without a robot or external tracker
(Fig.~\ref{fig:overview}, left). It does not reconstruct a trajectory:
translation is excluded because of drift. Rather, the recordings measure which
probe axis and sign the clinician uses at a given image offset. These
image--motion measurements parameterize the
Demonstration-Informed Image--Motion Controller (DIMC), an interpretable
controller that turns image features into a corrective probe-axis action
(Fig.~\ref{fig:overview}, middle). DIMC is combined with beam-axis force
regulation and an independent safety supervisor, so that the task-relevant
view is maintained while the contact condition changes
(Fig.~\ref{fig:overview}, right).

Unlike image--IMU methods for motion or volume reconstruction
\cite{luo2022monet,luo2023oscnet,wijkhuizen2026cnntransformer}, Sonologger
uses the synchronized signals as behavioral measurements for DIMC.

The bladder lumen is the motivating, dynamic-phantom target; the method is not
anatomy-specific beyond the segmentation model used to keep a structure in
view. Our contributions are:
\begin{enumerate}
\raggedright
\item \textbf{Robot-free image--motion demonstration acquisition.}\par
(Sec.~\ref{sec:acq}, \ref{sec:sync}) A probe-mounted inertial sensor,
single-clock synchronization, a measured latency correction and probe-frame
relative rotation labels, without robot-mediated data collection or external
cameras.
\item \textbf{IMU and relative-rotation validation with an explicit
measurement scope.}\par
(Sec.~\ref{sec:labels}) Six-axis orientation fusion
against FR5 kinematics, an explicit comparison with magnetometer-aided fusion,
and the explicit exclusion of translation from the controller because of
drift.
\item \textbf{Demonstration-Informed Image--Motion Controller (DIMC).}\par
(Sec.~\ref{sec:findings}, \ref{sec:dimc}) The corrective axis, its sign and
its activation threshold are measured from the demonstrations and encoded in
DIMC, an interpretable non-learned controller.
\item \textbf{Contact-adaptive safety-monitoring validation.}\par
(Sec.~\ref{sec:exp}) DIMC integrated with force regulation and safety
supervision is evaluated under syringe-induced dynamic contact changes and in
a blind, pose-paired dynamic bladder phantom test.
\end{enumerate}
We do not claim clinical validation, patient-specific clinical decision
autonomy, or safety certification. ``Contact-adaptive monitoring'' names the
intended systems role; here it is tested only as tissue- and contact-adaptive
behavior on a dynamic phantom. The robot is a probe holder that keeps a
clinician-selected view; the decisions that follow from the view remain the
clinician's.

\begin{figure*}[!t]
\centering
\makebox[\textwidth][l]{%
  \includegraphics[width=\textwidth]{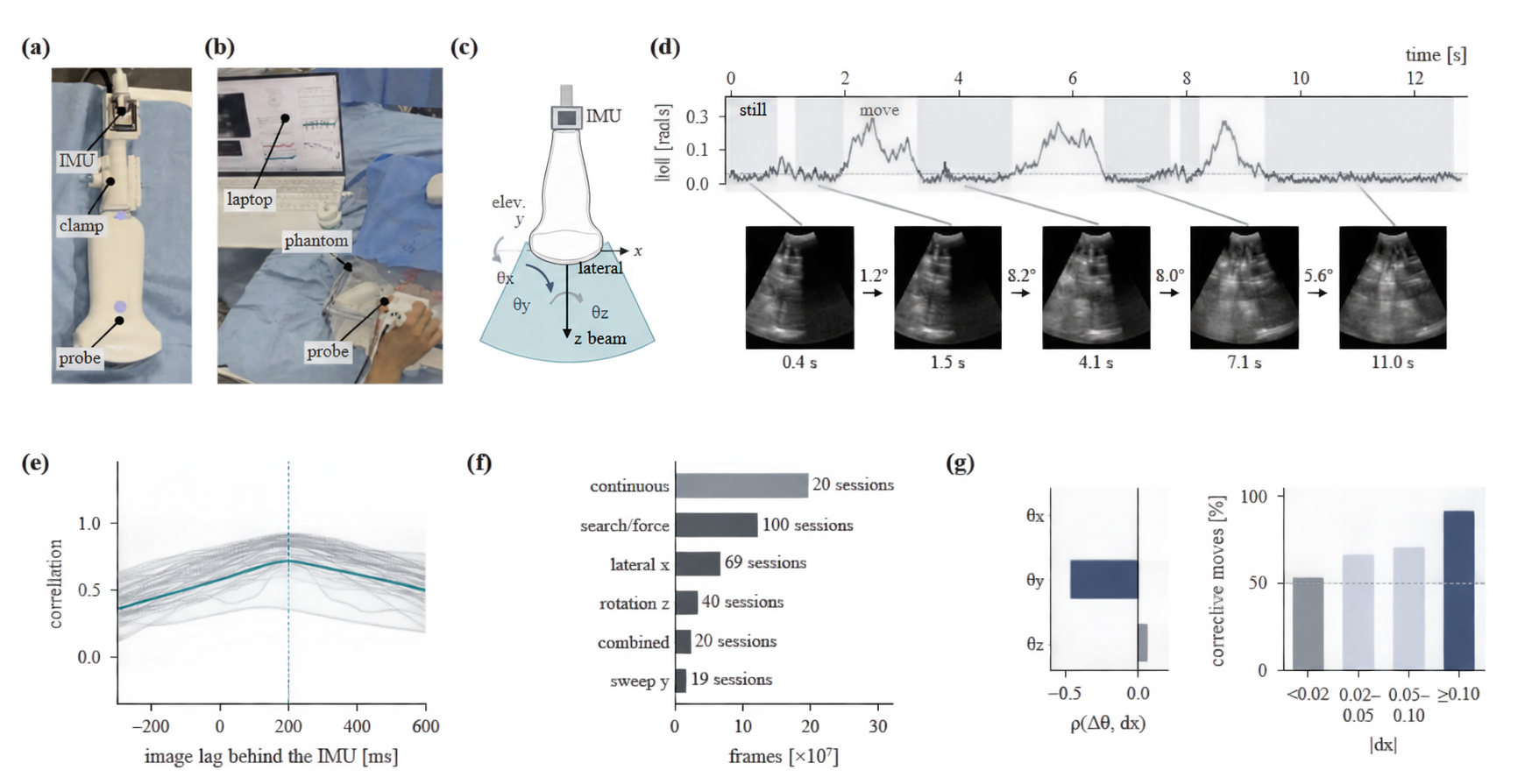}%
  \hspace{-\textwidth}%
  \rlap{%
    \raisebox{0.2648\textwidth}[0pt][0pt]{%
      \hspace*{0.1976\textwidth}%
      \setlength{\fboxsep}{1.1pt}%
      \raisebox{\depth}{\colorbox{white}{%
        \fontsize{6.2}{7.4}\selectfont\normalfont
        \begin{tabular}[b]{@{}c@{}}Sonologger\\(IMU + probe)\end{tabular}}}%
    }%
  }%
}
\caption{\textbf{Sonologger: robot-free image--motion acquisition.} (a)
BNO085 on an nRF52840 board clamped to the C10UR probe. (b) Freehand
acquisition on phantom A with a laptop; no robot, camera or tracker is
present. (c) Probe frame in which every relative-rotation label is expressed:
$x$ lateral, $y$ elevational (image-plane normal), $z$ beam; $\theta_y$ is the
in-plane rotation. (d) One lateral-search episode: gyroscope magnitude with
the still mask (grey), the sector image at each still and the relative
rotation between stills. (e) Effective image latency: correlation of image
change with the box-integrated gyroscope magnitude against lag, pooled over
226 sessions (thick) and for 40 sessions (thin); the peak is at
\SI{200}{\milli\second}. (f) Corpus composition: 268 sessions, \num{45902}
frames, \num{1818} move segments, \num{4389} chunks. (g) Behavioral
measurements from the demonstrations: only $\theta_y$ of (\ref{eq:label})
correlates with the lateral centroid offset before the move (Spearman
$\rho = \num{-0.46}$, $n = \num{3920}$ chunks), and the fraction of corrective
moves rises from chance (dashed) inside the dead band to \SI{92}{\percent} at
$|dx| \ge \num{0.10}$ ($n = 518$, 525, 649 and 536 per bin).}
\label{fig:sono}
\end{figure*}

\section{Related Works}

\subsection{Robotic Ultrasound Scanning and Demonstration-Based Control}

Robotic ultrasound and its control are surveyed
in~\cite{priester2013robotic,vonhaxthausen2021medical,li2021overview,jiang2023robotic}. Many systems execute a protocol:
they press along a planned trajectory registered to a preoperative
volume~\cite{hennersperger2017towards}, normalize the probe to the surface
from the confidence map and the measured force~\cite{jiang2020automatic},
compensate the motion of the scanned surface during a
sweep~\cite{jiang2021motion}, or steer towards standard planes with
reinforcement learning~\cite{li2021autonomous,bi2022vesnet}. Learning from
demonstration normally assumes the demonstration is recorded on the
learner~\cite{argall2009survey,ravichandar2020recent}, and action-chunking
imitation methods~\cite{zhao2023act} inherit that trajectory-centric assumption. Learned
image-guidance policies commonly formulate acquisition as reaching a
pre-specified standard plane or image-quality target~\cite{li2021autonomous,bi2022vesnet,hsu2026active}.
Robot-free hand-held interfaces have also enabled human demonstration collection
for general manipulation~\cite{chi2024umi}; Sonologger instead measures
ultrasound image--motion relations rather than transferring trajectories.
The present work runs in the other
direction. It does not learn a policy from robot-collected sweeps; it measures,
from robot-free human image--motion data, the interpretable action parameters
of a monitoring controller, which axis is corrected, in which sign and from
what offset, and leaves the plant-specific gain to the robot.

\subsection{Freehand Image--IMU Ultrasound Reconstruction and Robot-Free Demonstrations}

Freehand 3D reconstruction uses external tracking~\cite{prager1999stradx},
speckle decorrelation~\cite{housden2007sensorless}, image--IMU fusion
\cite{prevost2018freehand}, or sensorless learning-based reconstruction
\cite{luo2023recon}. MoNet fuses image features with a single IMU
\cite{luo2022monet}; OSCNet uses multiple IMUs and self-consistency
\cite{luo2023oscnet}; and a CNN--Transformer method combines images and IMU
orientation~\cite{wijkhuizen2026cnntransformer}. RecON instead learns
sensorless reconstruction from ultrasound sequences~\cite{luo2023recon}.
These methods target motion or volume reconstruction, not the image--motion
behavioral measurements used here.

Unlike image--IMU approaches that estimate inter-frame transformations or
reconstruct 3D volumes, our framework does not pursue global probe
localization or volumetric reconstruction. We use synchronized ultrasound and
probe-relative rotation recordings as behavioral measurements: they identify
which probe axis a clinician uses to correct an image displacement, in which
direction, and beyond which image-error range. These measurements parameterize
an interpretable Demonstration-Informed Image--Motion Controller rather than a
learned motion-reconstruction model. The robot-free collection process avoids
robot-mediated demonstrations; it does not claim robot-independent absolute
action labels or trajectory transfer. Zero-velocity bracketing of inertial
integration is borrowed from pedestrian navigation~\cite{foxlin2005pedestrian,skog2010zerovelocity},
and the failure of magnetometer-aided fusion near ferromagnetic structure is
the one reported for motion laboratories~\cite{devries2009magnetic}.

\subsection{Contact Regulation and Dynamic Ultrasound Monitoring}

Contact is regulated by hybrid force--position control for assisted
scanning~\cite{gilbertson2015force}, while confidence- and image-guided
servoing adapt probe pose~\cite{chatelain2017confidence,jiang2020automatic}.
Bladder ultrasound segmentation---including anterior-wall and low-compute
region segmentation---is well covered~\cite{saini2026bwsnet,song2023memory}, and reinforcement learning has
been used to steer a bladder scan towards a better view~\cite{hsu2026active};
our perception stage follows the standard U-Net recipe~\cite{ronneberger2015unet}
pretrained on a public pelvic floor corpus~\cite{solismartin2026pfus1}; dynamic
CNN-based pelvic-floor organ identification has likewise been demonstrated in
transperineal ultrasound~\cite{garciamejido2024applicability};
the dynamic phantom follows a published tissue-mimicking
recipe~\cite{fernandez2024phantom}. These systems servo towards a view or hold
a force. The role considered here is different: a clinician chooses the view
needed to observe tissue while a separate primary procedure proceeds, and the
robot maintains that view while bounding probe--tissue interaction. Our
contribution is the combination of robot-free demonstration-derived
image--motion parameters, beam-axis force regulation, and an independent
safety supervisor, evaluated only on task-relevant view maintenance under
changing contact conditions in a phantom.

\section{Method}
\label{sec:method}

\subsection{Sonologger and Robot-Free Image--Motion Acquisition}
\label{sec:acq}

A BNO085 inertial unit on an nRF52840 board is clamped to a Konted C10UR
wireless convex probe in a printed shell (Fig.~\ref{fig:sono}a). Accelerometer
and gyroscope data are fused at \SI{200}{\hertz} with a magnetometer-free
gradient-descent filter~\cite{madgwick2011estimation}; ultrasound frames arrive
at \SI{10}{\hertz}. Both streams are logged on one laptop clock (Fig.~\ref{fig:sono}b).
Thus acquisition adds only the clamp, sensor and laptop---no robot, external
camera or optical tracker.

\subsection{Synchronization, Latency Correction, and Still--Move--Still Segmentation}
\label{sec:sync}

The effective ultrasound delay $\tau$ is estimated by maximizing the
correlation between frame-to-frame image change and time-shifted gyroscope
magnitude. It is \SI{200}{\milli\second} when pooled over 226 sessions
(Fig.~\ref{fig:sono}e) and is removed once in dataset construction.

Still--move--still segmentation uses thresholds on gyroscope and acceleration
variation in a \SI{0.30}{\second} window. Each accepted still-to-still interval
(\SIrange{0.15}{8.0}{\second}) yields its relative-rotation label and the 20
preceding ultrasound frames; labels are provided as \SI{1.6}{\second} chunks.

Two days of scanning by one operator on phantom A yield 268 sessions,
\num{45902} frames, \num{1818} move segments and \num{4389} chunks
(Fig.~\ref{fig:sono}f). Splits are session-wise and therefore assess
reproducibility, not generalization.

\subsection{Probe-Frame Relative Rotation and IMU Quality Validation}
\label{sec:labels}

Each move is represented by its probe-frame rotation between the bracketing
stills. In the frame of Fig.~\ref{fig:sono}c, $x$ is lateral, $y$ elevational
(the image-plane normal) and $z$ the beam; $\theta_y$ is the in-plane rotation
used to move image content laterally. With a once-per-mount sensor-to-probe
calibration $R_{PS}$, the label for anchors $a$ and $b$ is
\begin{equation}
\boldsymbol\theta_{ab} = \log\!\big(R_{PS}\, R_{ES}(t_a)^{\top} R_{ES}(t_b)\, R_{PS}^{\top}\big)^{\vee}
= (\theta_x, \theta_y, \theta_z),
\label{eq:label}
\end{equation}
the relative-rotation vector in degrees. This removes the unobservable global
heading of a six-axis filter and makes the axes consistent across moves. The
calibration fixes axis assignment; the lateral sign is set from observed image
motion and confirmed by the robot self-probe.

Table~\ref{tab:imu} validates the measurement scope against FR5 kinematics.
Magnetometer-free fusion gives \SI{2.52}{\degree} geodesic RMS on the
teleoperated recording, whereas the on-chip magnetometer-aided estimate is
substantially degraded near the steel arm. We therefore use only short
relative rotations. Translation accumulates \SI{6.3}{\milli\metre} error in
\SI{5}{\second} and is recorded but never supplied to DIMC; this is a
relative-orientation measurement, not trajectory reconstruction.

\begin{table}[H]
\bodytablecaption{Probe-mounted IMU validation against FR5 kinematics and scope
of use. Same \SI{84}{\second} teleoperated recording (\num{16757} paired
samples), same hand-eye alignment (seven static poses, \SI{0.81}{\degree}
RMS residual).}
\label{tab:imu}
\vspace{2pt}
\centering
\footnotesize\normalfont
\setlength{\tabcolsep}{3pt}
\begin{tabular}{>{\raggedright\arraybackslash}p{2.0cm}>{\raggedright\arraybackslash}p{1.9cm}>{\raggedright\arraybackslash}p{1.5cm}>{\raggedright\arraybackslash}p{2.3cm}}
\toprule
Quantity / configuration & RMS / 95th pct. & Drift & Role in DIMC \\
\midrule
Six-axis fusion, no magnetometer & \SI{2.52}{\degree} / \SI{3.61}{\degree} & \SI{0.23}{\degree\per\minute} & Used for short relative rotation labels \\
Nine-axis on-chip rotation vector & \SI{12.55}{\degree} / \SI{18.91}{\degree} & \SI{8.29}{\degree\per\minute} & Rejected: magnetic distortion near the robot \\
IMU translation\newline integration & \SI{6.3}{\milli\metre}\newline at \SI{5}{\second} & accum. & Recorded; excluded from DIMC \\
\bottomrule
\end{tabular}
\end{table}

\subsection{Image--Motion Measurements From Freehand Demonstrations}
\label{sec:findings}

\textbf{Image features.} Polar frames are converted to a sector image and
segmented by a U-Net pretrained on a public pelvic-floor corpus and fine-tuned
on the demonstration and dynamic-phantom data. Frames with insufficient
acoustic coupling or implausible masks are physically gated out. DIMC then
uses the gated mask area, lateral centroid offset $dx$ (positive rightward),
and largest-component fraction. A move is \emph{corrective} when its
$\theta_y$ sign reduces $|dx|$ under the calibrated plant sign. Thus the
calibration relates probe rotation to image motion, while the demonstrations
identify the axis, sign and offset at which the clinician chooses to correct.

Across the target-present chunks, lateral offset is associated only with the
in-plane rotation $\theta_y$ (Spearman $\rho=\num{-0.463}$; Fig.~\ref{fig:sono}g),
not with out-of-plane or beam-axis rotation. Corrective moves rise from near
chance inside the small-offset region to \SI{91.6}{\percent} at large offsets.
DIMC therefore uses $\theta_y$, its demonstrated sign, and a \num{0.03}
dead band (about \SI{8}{\milli\metre}) at the edge of that region. The detailed
bin counts and axis-wise statistics are shown in Fig.~\ref{fig:sono}g.

The demonstration and the robot self-probe agree on the corrective sign, but
not its magnitude: the robot-specific gain is consequently measured by the
\SI{60}{\second} self-probe, whereas the axis, sign and activation threshold
are transferred from the demonstrations.

\subsection{Demonstration-Informed Image--Motion Controller}
\label{sec:dimc}

\begin{figure}[t]
\centering
\includegraphics[width=\linewidth]{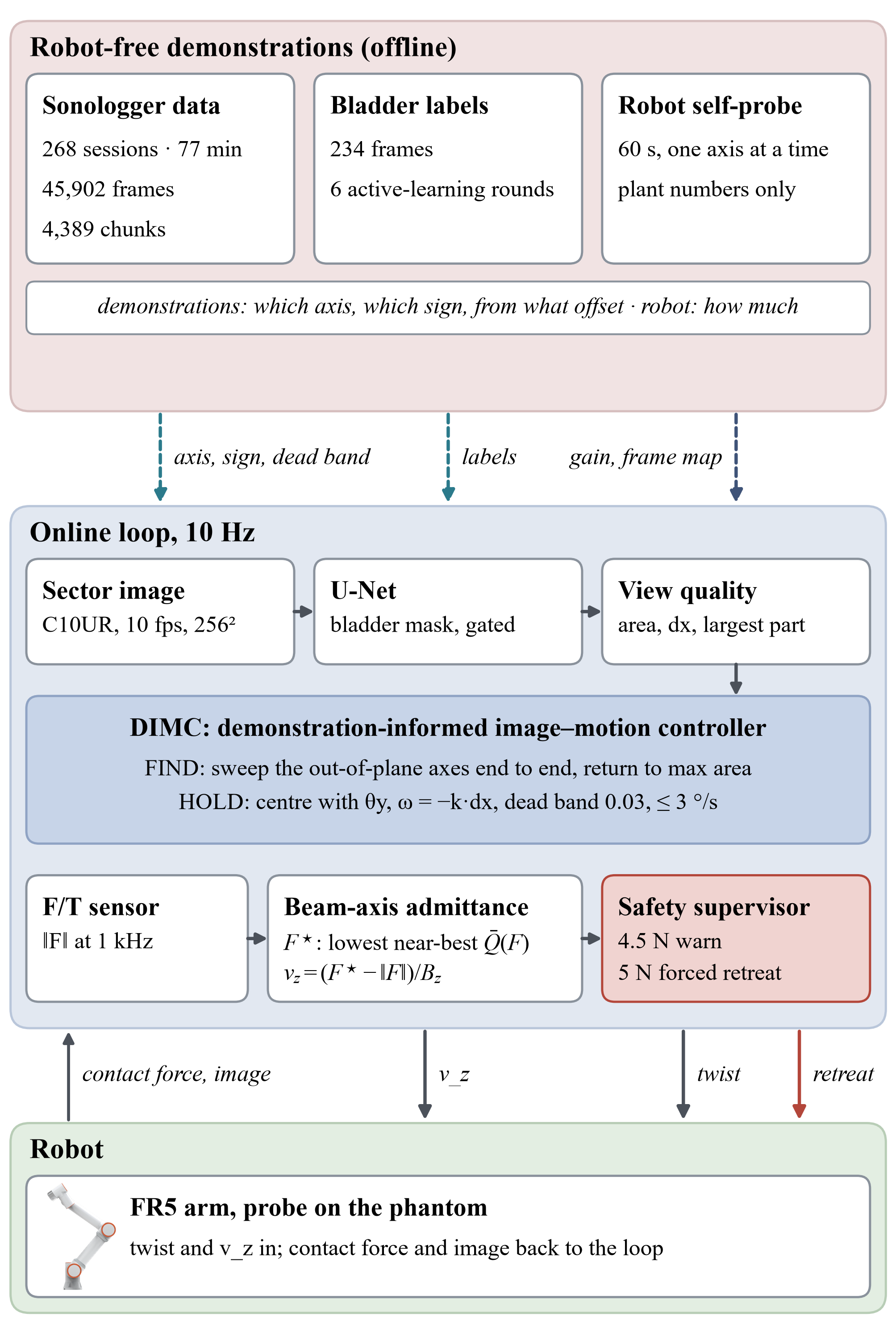}
\caption{\textbf{From robot-free demonstrations to DIMC and the force and
safety stack.} Offline (dashed) the demonstrations supply the behavioral
parameters of the Demonstration-Informed Image--Motion Controller (DIMC),
which axis is corrected, in which sign and from what offset, plus the U-Net
labels; a \SI{60}{\second} robot self-probe supplies the gain magnitude, the
image responses, the frame map and the safe search ranges. Online (solid,
\SI{10}{\hertz}) the sector image is segmented, reduced to area, centroid and
largest-component features, and turned into a twist by the \emph{Find} or \emph{Hold} mode
of DIMC; the beam axis is owned by the admittance loop and the independent
safety supervisor.}
\label{fig:controller}
\end{figure}

The Demonstration-Informed Image--Motion Controller (DIMC) is an
interpretable, non-learned controller whose corrective probe axis, action
sign, and activation threshold are measured from robot-free image--motion
demonstrations. It is neither a neural, learned or imitation policy nor a
hand-tuned controller: its behavioral structure is fixed, but the corrective
axis, the sign and the dead band are behavioral measurements taken from the
human demonstrations of Sec.~\ref{sec:findings}. Fig.~\ref{fig:controller}
shows its three stages. In the \emph{offline stage} the robot-free corpus
supplies the corrective probe axis ($\theta_y$), its sign, the dead band
(\num{0.03}) and the perception labels. A \emph{robot self-probe} of
\SI{60}{\second}, one axis at a time, supplies the robot-specific quantities:
the gain magnitude (\SI{+0.01913}{\per\degree}), the image response of each
axis (the in-plane rotation changes the area by \SI{+0.00206}{\per\degree},
the out-of-plane tilt by \SI{-0.00189}{\per\degree}), the frame map and the
safe search ranges. In the \emph{online stage}, at \SI{10}{\hertz}, the gated
U-Net output is reduced to area, centroid offset and the largest-component
fraction, and one of two operational modes issues a twist.

\textbf{Find} performs a safe search when the task-relevant structure is not
visible: it sweeps the out-of-plane axes identified by the self-probe end to
end (\SI{\pm 12}{\degree} or \SI{\pm 25}{\milli\metre}, widened on each
unsuccessful pass up to \SI{30}{\degree} or \SI{60}{\milli\metre}) and returns
to the pose of maximum mask area; it is a sweep rather than hill climbing
because the area of an unseen structure is zero and has no gradient.
\textbf{Hold} maintains the view by in-plane corrective adaptation once the
structure is visible: proportional centering on the in-plane rotation,
$\omega_y = -k\,dx/g$ with the self-probe gain $g$ and
$k = \SI{0.4}{\per\second}$, the demonstration-derived dead band of
\num{0.03} and a rate limit of \SI{3}{\degree\per\second}; because the
in-plane rotation also changes the area, the cumulative correction is capped
and, if the area falls below half of its best value, DIMC returns to the best
pose and re-enters \emph{Find}. Transitions use hysteresis on the area. The beam
axis is never touched by either mode; it belongs to the force loop of
Sec.~\ref{sec:force}. The axis, the sign and the dead band come from the
demonstrations; the gain magnitude, the image responses and the axes \emph{Find}
sweeps come from the self-probe; the sweep ranges, the rate limit, the caps
and the hysteresis were set inside the safety envelope, and all were frozen
before the blind test.

\subsection{Force Regulation and Safety Supervision}
\label{sec:force}

\begin{figure*}[tp]
\centering
\includegraphics[width=\textwidth]{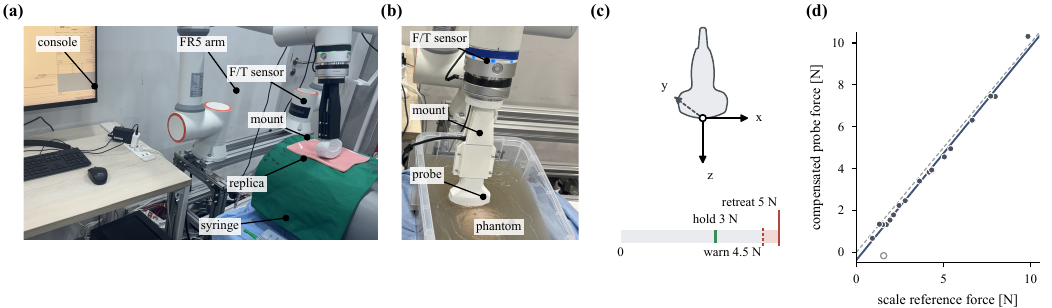}
\caption{\textbf{Force regulation and safety supervision hardware.} (a) FR5
arm, PX6D force/torque sensor, printed mount and a probe replica on the force
bench, with the console and the \SI{500}{\milli\litre} syringe. (b) The same
stack with the C10UR probe on the gel phantom. (c) Probe frame ($x$ lateral,
$y$ elevational, $z$ beam) and the safety supervisor's envelope on
$\lVert\mathbf F\rVert$: \SI{3}{\newton} hold target, \SI{4.5}{\newton}
warning, \SI{5}{\newton} forced retreat. (d) Compensated probe force against a
laboratory scale over \SIrange{0.92}{9.85}{\newton} ($n = 19$; slope \num{1.019},
one invalid trial open; dashed: identity, solid: fit).}
\label{fig:hardware}
\end{figure*}

The FR5 carries a compensated six-axis force/torque sensor, printed mount and
C10UR probe (Fig.~\ref{fig:hardware}a, b); a working-pose tare is acquired
before each run. The beam axis $z$ is owned by an admittance regulator,
$v_z = (F^\star - \lVert \mathbf F \rVert)/B_z$, while DIMC controls only the
image-motion axes. $F^\star$ is not universally \SI{3}{\newton}: after a
raw-image coupling gate, it is the lowest safe force with near-best image
quality, $F^\star=\min\{F:\bar Q(F)\ge\bar Q_{\max}-\delta_Q\}$, where
$\delta_Q$ is noise-aware. The fixed \SIrange{0.5}{4.0}{\newton} holds isolate
regulation and safety, rather than validate online force search. The independent
supervisor (Fig.~\ref{fig:hardware}c) blocks rising commands above
\SI{4.5}{\newton}, commands a $-z$ retreat at \SI{5}{\newton}, and retreats on
command loss. Calibration and regulator details are in Fig.~\ref{fig:hardware}d
and Sec.~\ref{sec:exp_force}.

\section{Experiments}
\label{sec:exp}

\subsection{Experimental Platforms and Dynamic Contact Conditions}

\begin{figure*}[tp]
\centering
\includegraphics[width=\textwidth]{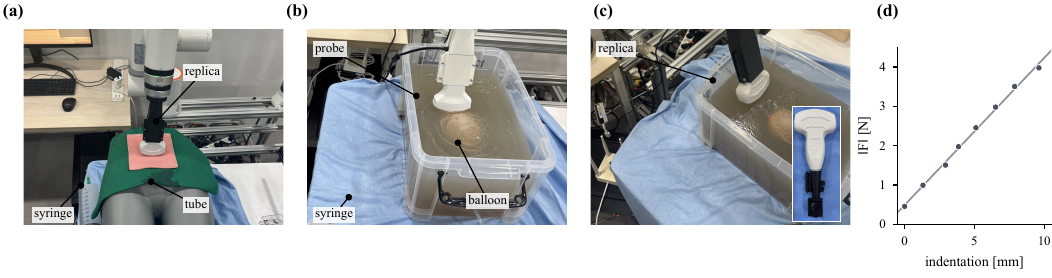}
\caption{\textbf{Platforms for dynamic contact conditions.} (a) Force bench: a
torso mannequin with an inflatable tube under the abdominal pad, driven by a
\SI{500}{\milli\litre} syringe, and a 3D-scanned, printed replica of the
clinical C10UR probe. (b) Dynamic
bladder phantom: an agar-gelatin block with a double water-balloon lumen
filled and emptied through a syringe line, scanned by the robot-held C10UR.
(c) Contact-stiffness measurement of the gel with the same replica (inset). (d)
Equilibrium force against indentation on the mannequin bench from the force
holds, $k = \SI{0.374}{\newton\per\milli\metre}$, $R^2 = \num{0.997}$.}
\label{fig:benches}
\end{figure*}

Three platforms answer distinct questions (Fig.~\ref{fig:benches}). Phantom A
provides the freehand Sonologger demonstrations. A syringe-driven mannequin
force bench tests regulation and forced retreat under surface displacement.
The dynamic bladder phantom has a syringe-filled lumen whose volume and probe
contact change together; it is used for the blind view-acquisition and
hydro-distension tests. These are phantom demonstrations of contact-adaptive
monitoring, not clinical validation.

\subsection{Force Regulation and Safety Evaluation}
\label{sec:exp_force}

We tested eight force targets from \SIrange{0.5}{4.0}{\newton} in two
sessions (16 holds) and compared syringe injections and withdrawals with the
admittance loop on versus off (70 events per condition). We measure hold
error, recovery after displacement, probe travel, and supervisor response.

\begin{figure*}[!tp]
\centering
\includegraphics[width=0.86\textwidth]{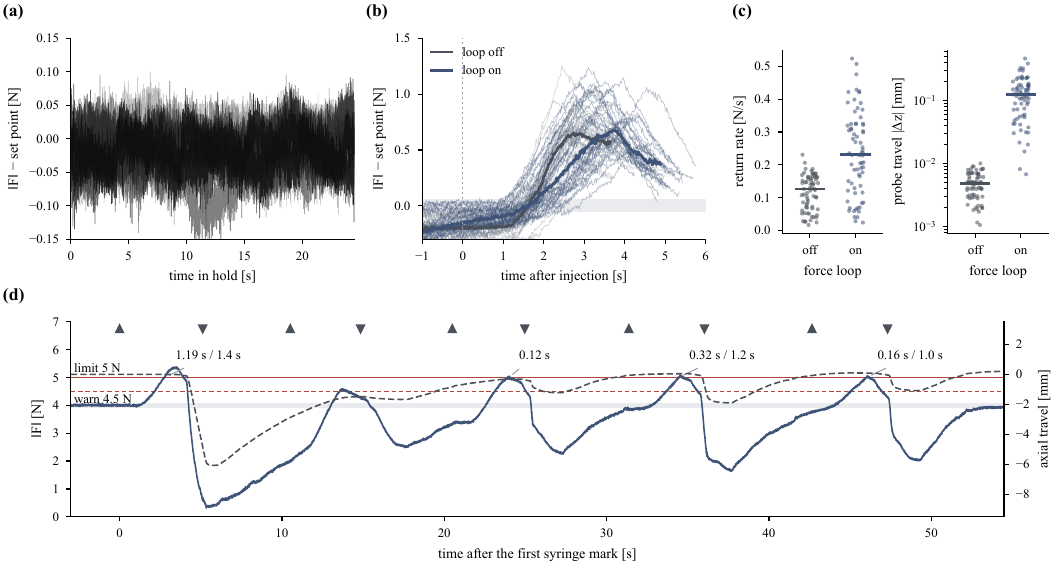}
{\setlength{\abovecaptionskip}{-1pt}%
\caption{\textbf{Force regulation and safety response during
syringe-induced surface displacement.} Force holds, loop-on/off injection
responses, and the \SI{4.0}{\newton} safety-retreat run are shown.}
\label{fig:force}%
}
\end{figure*}

Across the 16 holds the standard deviation of the held force is
\SI{0.028}{\newton} on average (\SIrange{0.022}{0.049}{\newton} per set
point), the mean error of every hold lies inside the \SI{\pm 0.05}{\newton}
dead band, and the two sessions agree to within \SI{0.047}{\newton}
(Fig.~\ref{fig:force}a, Table~\ref{tab:force}). Under the syringe-induced
surface displacement the loop does not reduce the peak of an injection
(\SI{0.83}{\newton} against \SI{0.74}{\newton} with the loop off) but changes
everything after it: the probe travels \SI{0.13}{\milli\metre} per event
against \SI{0.005}{\milli\metre}, and the post-peak return rate rises from
\SI{0.13}{\newton\per\second} to \SI{0.23}{\newton\per\second}
($p < \num{0.001}$, 70 events per arm; Fig.~\ref{fig:force}b, c). The peak
is set by how fast the surface rises against the \SI{12}{\second} time
constant, and the bench is deliberately faster than the task: the syringe
delivers \SI{500}{\milli\litre} within a few seconds, whereas the volume of a
monitored organ under irrigation changes over tens of seconds to minutes. The
safety supervisor was exercised rather than assumed: at the \SI{4.0}{\newton}
set point four of five injections drove the force past \SI{5}{\newton}
(Fig.~\ref{fig:force}d); each time the retreat fired, the force stayed above
\SI{5}{\newton} for \SIrange{0.12}{1.19}{\second} and was back below the
\SI{4.5}{\newton} warning within \SIrange{1.0}{1.4}{\second}, and at every
lower set point the peak stayed below the warning line.

\begin{table}[!hb]
\bodytablecaption{Force hold on the mannequin bench. Pooled over two
independently calibrated sessions, \SI{30}{\second} per hold, first
\SI{3}{\second} discarded, \SI{1}{\kilo\hertz} sampling.}
\label{tab:force}
\vspace{4pt}
\centering
\footnotesize\normalfont
\begin{tabular}{rrrrr}
\toprule
Target [N] & SD [N] & Mean error [N] & In band [\%] & Peak [N] \\
\midrule
0.5 & 0.023 & $-0.017$ & 88.9 & 0.60 \\
1.0 & 0.022 & $-0.013$ & 83.9 & 1.10 \\
1.5 & 0.022 & $-0.012$ & 94.1 & 1.57 \\
2.0 & 0.026 & $-0.016$ & 89.9 & 2.09 \\
2.5 & 0.029 & $-0.031$ & 72.9 & 2.59 \\
3.0 & 0.026 & $-0.015$ & 89.7 & 3.10 \\
3.5 & 0.025 & $-0.008$ & 93.7 & 3.59 \\
4.0 & 0.049 & $-0.025$ & 82.9 & 4.36 \\
\bottomrule
\end{tabular}
\end{table}

\begin{figure*}[!t]
\centering
\includegraphics[width=0.82\textwidth]{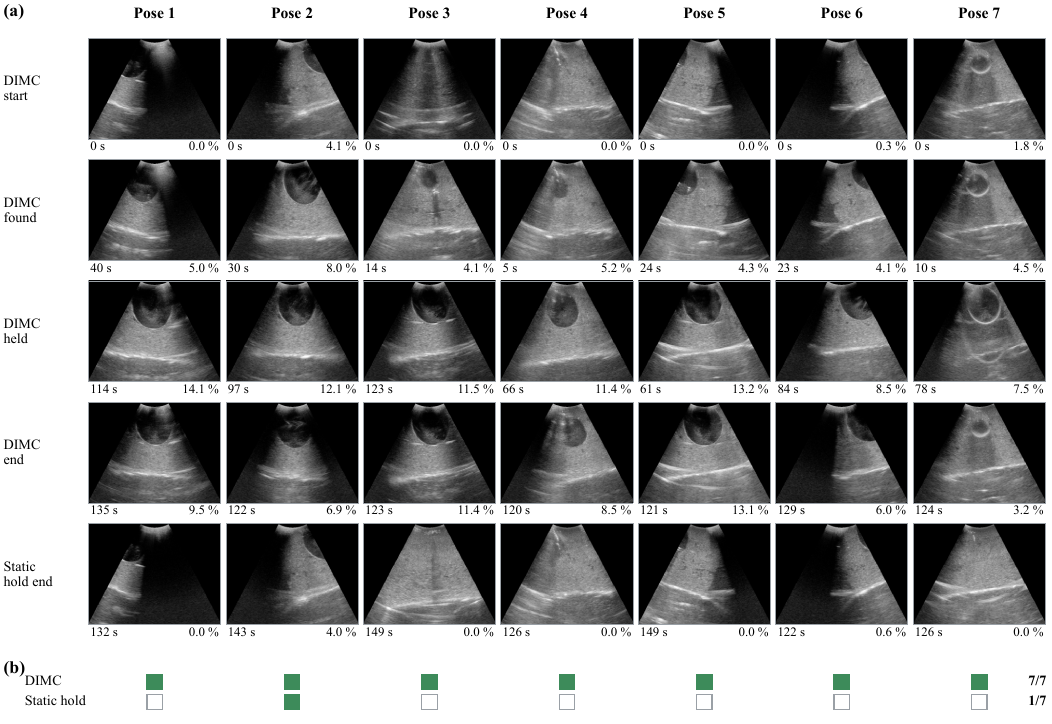}
\caption{\textbf{Blind task-relevant view acquisition and maintenance.}
Seven DIMC episodes and static-hold outcomes; DIMC maintained the view in 7/7
poses versus 1/7 for static hold.}
\label{fig:qual}
\end{figure*}

\subsection{Blind Task-Relevant View Acquisition and Maintenance}
\label{sec:exp_main}

The blind test compares DIMC with a static hold from seven randomized,
operator-blinded starting poses on the dynamic phantom (\SI{120}{\second} of closed-loop
control per episode, scored to the end of the recorded state stream,
\SI{123}{\second} for pose~1). Times are measured from the start of
control. Both conditions use the same force and safety stack; static hold
issues no image command. Success is a task-relevant view maintained for
\SI{3}{\second}, using pre-frozen mask-area, component and centering criteria.
The analysis is pose-paired.

Fig.~\ref{fig:qual} shows every DIMC episode from its blind start to its held
view, and Table~\ref{tab:main} reports the per-pose outcome. DIMC reached and held a
task-relevant view in 7 of 7 poses (Wilson 95\% CI \SI{64.6}{\percent} to
\SI{100}{\percent}), against 1 of 7 for the static hold (\SI{2.6}{\percent}
to \SI{51.3}{\percent}); all six discordant pairs favor DIMC (McNemar exact
$p = \num{0.031}$). The one static-hold success, pose 2, started with the
lumen already visible (initial area \num{0.029}; the start gate was not
enforced and the episode is kept). DIMC reached its first task-relevant view
after a median of \SI{13.0}{\second} (range \SIrange{7.8}{50.7}{\second}) and
then held one for \SIrange{73}{112}{\second} of the remaining episode.

{\setlength{\intextsep}{4pt}%
\begin{table}[H]
\bodytablecaption{\textbf{Blind pose-paired test on the dynamic phantom.} Time to the
first task-relevant view (held for \SI{3}{\second}; ``no'' if never reached)
per pose and condition.}
\label{tab:main}
\vspace{4pt}
\centering
\begin{tabular}{rrr}
\toprule
Pose & DIMC [s] & Static hold [s] \\
\midrule
1 & 50.7 & no   \\
2 & 10.9 & 35.4 \\
3 & 13.0 & no   \\
4 & 7.8  & no   \\
5 & 25.4 & no   \\
6 & 16.4 & no   \\
7 & 9.3  & no   \\
\midrule
Found & 7/7 & 1/7 \\
\bottomrule
\end{tabular}
\end{table}}

\subsection{View Maintenance Under Hydro-Distension}
\label{sec:exp_hydro}
\begin{figure}[!t]
\centering
\includegraphics[width=0.82\linewidth]{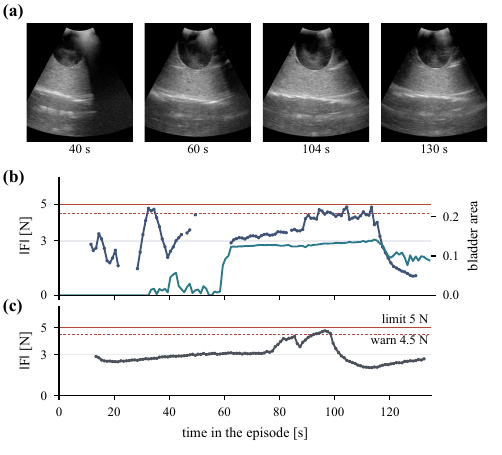}
\caption{\textbf{Hydro-distension view maintenance.} DIMC retains the target
through the force peak; static hold lacks image-centering adaptation.}
\label{fig:hydro}
\end{figure}

During pose 1, syringe filling and withdrawal changed lumen volume and probe
contact while each condition ran; no episode was aborted by the supervisor.
DIMC found the lumen after \SI{50.7}{\second} of control
(Table~\ref{tab:main}; \SI{62}{\second} on the recording clock of
Fig.~\ref{fig:hydro}) and retained it through filling
and supervisor retreat: force peaked at \SI{5.07}{\newton}, but the mask
remained above the task threshold. Static hold remained below the limit (peak
\SI{4.77}{\newton}) but could not provide image-centering adaptation.


\raggedbottom
\section{Discussion and Conclusion}
\label{sec:limits}
Sonologger collects robot-free image--motion demonstrations from the
clinician-operated probe with a clamp, a sensor and a laptop, adding nothing
to the clinician's routine and bringing no robot near the subject. The
demonstrations are not trajectory-imitation data; they are behavioral
measurements that identify which probe axis is corrected, in which direction,
and from what image offset. These measurements are used to design an
image-conditioned corrective probe action. DIMC converts these measurements into an interpretable, non-learned
probe-axis adaptation whose \emph{Find} and \emph{Hold} modes search for and
maintain the task-relevant view. Rather than reproducing a fixed scan or
expert trajectory, it gives the robot an operational representation of
ultrasound context: which image change warrants correction, along which probe
axis, and in which direction.

Together with beam-axis force regulation and an independent safety supervisor,
this representation lets an intraoperative robotic assistant perform a
task-specific supporting action within a defined safety boundary. DIMC
acquired and held a task-relevant view in 7 of 7 blind poses, versus 1 of 7
for static hold, and retained the view during syringe-driven hydro-distension
while the supervisor bounded the force excursion. The intended role is not
autonomous scanning for its own sake, but continuous, contact-adaptive
observation that supports a primary procedure as tissue state, contact and
exposure change.

After workflow-specific validation, the same acquisition and
parameterization can support organ or lesion monitoring, tissue-plane
observation and intermittent ultrasound guidance across surgical specialties;
bladder monitoring during urologic surgery is one motivating case. This is a
single-operator phantom study, not a clinical validation: no patient data,
clinical deployment, absolute translation or global trajectory are claimed.
Future work will collect multi-operator and approved observational clinical
data, establish workflow-specific parameters and prospectively validate them
in intended intraoperative settings. Code and the main-test records are
available at \url{https://github.com/suung23/FR5-for-RUS}; the demonstration
corpus is available from the authors.

\section*{Acknowledgment}
This work was supported by the Institute of Information \& Communications
Technology Planning \& Evaluation (IITP) grant funded by the Korea government
(MSIT) (IITP-2026-RS-2026-25527579, Development of Synthetic Data Extension and
Augmentation Technology for Medical Procedure Action Data), and by the
Technology Development Program (RS-2025-02306259) funded by the Ministry of SMEs
and Startups (MSS, Korea).

\setlength{\columnsep}{12pt}
\bibliographystyle{IEEEtran}
\bibliography{refs}
\end{document}